\documentclass[runningheads]{llncs}
\usepackage{graphicx}

\usepackage{tikz}
\usepackage{comment}
\usepackage{amsmath,amssymb} 
\usepackage{color}

\usepackage[accsupp]{axessibility}  

\usepackage{graphicx}
\usepackage{amsmath}
\usepackage{amssymb}
\usepackage{booktabs}

\usepackage{placeins}
\usepackage{adjustbox}
\graphicspath{{./images/plots} }
\usepackage{float}

\usepackage[pagebackref,breaklinks,colorlinks]{hyperref}

\usepackage{hyperref}
\usepackage{orcidlink}

\begin{document}
\pagestyle{headings}
\mainmatter
\def\ECCVSubNumber{9}  

\title{SOMPT22: A Surveillance Oriented Multi-Pedestrian Tracking Dataset} 

\titlerunning{SOMPT22 Dataset}
%

\author{Fatih Emre Simsek\inst{1,2}\orcidlink{0000-0003-3362-5747} \index{Simsek, Fatih Emre} \and
Cevahir Cigla\inst{1}\orcidlink{0000-0003-3246-3176} \and
Koray Kayabol\inst{2}\orcidlink{0000-0003-0053-2800}}
\authorrunning{F. E. Simsek et al.}
%
\institute{Aselsan INC, Ankara, TURKEY \\
\email{\{fesimsek,ccigla\}@aselsan.com.tr} \and
Gebze Technical University, Kocaeli, TURKEY \\
\email{koray.kayabol@gtu.edu.tr}
}
\maketitle

\begin{figure}
  \centering
  \includegraphics[width=\linewidth]{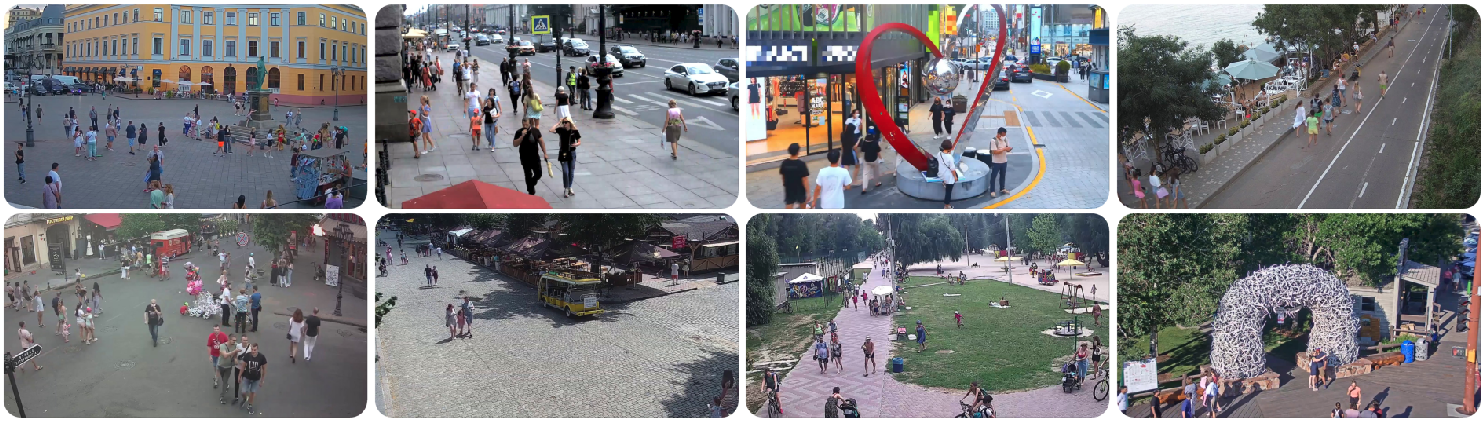}
\end{figure}

\begin{abstract}
Multi-object tracking (MOT) has been dominated by the use of track by detection approaches due to the success of convolutional neural networks (CNNs) on detection in the last decade. As the datasets and bench-marking sites are published, research direction has shifted towards yielding best accuracy on generic scenarios including re-identification (reID) of objects while tracking. In this study, we narrow the scope of MOT for surveillance by providing a dedicated dataset of pedestrians and focus on in-depth analyses of well performing multi-object trackers to observe the weak and strong sides of state-of-the-art (SOTA) techniques for real-world applications. For this purpose, we introduce \textit{SOMPT22} dataset; a new set for multi person tracking with annotated short videos captured from static cameras located on poles with 6-8 meters in height positioned for city surveillance. This provides a more focused and specific benchmarking of MOT for outdoor surveillance compared to public MOT datasets. We analyze MOT trackers classified as one-shot and two-stage with respect to the way of use of detection and reID networks on this new dataset. The experimental results of our new dataset indicate that SOTA is still far from high efficiency, and single-shot trackers are good candidates to unify fast execution and accuracy with competitive performance. 
The dataset will be available at: \href{https://sompt22.github.io}{sompt22.github.io}

\end{abstract}

\section{Introduction}
Multi-object tracking (MOT) is a popular computer vision problem that focuses on tracking objects and extracting their trajectories under different scenarios that are later used for various purposes. The use cases of MOT can be various including but not limited to autonomous cars, video surveillance, missile systems, and sports. The output of MOT could be predicting the next positions of the objects, prevent collusion or extract some statistics about the scene and object behaviours. Detection is the first and perhaps the most crucial stage in MOT to identify the object or target and define the purpose of the track. There are different well-defined types of objects such as pedestrians, faces, vehicles, animals, air-crafts, blood cells, stars or any change as a result of motion with respect to background that are subject to the need of tracking in several scenarios. In real world, monitored objects can alter visibility within time due to environmental conditions, physical properties of the object/motion or crowdedness. The change in visibility as well as existence of multiple objects yield occlusion, object/background similarities, interactions between instances. These effects combined with environmental factors such as illumination change, introduce common challenges and difficulties in MOT literature both in detection and tracking stages. 

In recent years, various algorithms and approaches have been proposed via CNNs to attack aforementioned problems. The research on MOT has also gained popularity with the appearance of the first MOTChallenge \cite{motchallenge} in 2014. This challenge contains annotated videos, with object (pedestrian) bounding boxes among consecutive frames that are captured through moving or static cameras. In the MOT challenge, the variety of data is large, including eye-level videos captured on moving vehicles, high-level moving, or stationary surveillance cameras at various heights.
This variation of the data limits the understanding of the capabilities of state-of-the-art (SOTA) techniques such that the positions of the cameras and the types of scene objects change the scenarios, as mentioned previously. These challenges require MOT approaches to be generic and capable of tracking vehicles or pedestrians under varying conditions. However, MOT techniques, as well as any other technique, should be optimized per scenario to get maximum efficiency and low false alarms for real-world usage.

In this paper, we focus on MOT challenge on static and single cameras positioned for the sake of surveillance of pedestrians at around 6-8 meters in height. In this manner, we create a new dataset and analyze the well-known MOT algorithms in this dataset to understand the performance of SOTA for surveillance. This approach is expected to optimize MOT algorithms by further analyzing motion behaviour and object variations for semi-crowded scenes in the limited scope. Public MOT datasets challenge the detection and tracking performances of MOT algorithms by increasing the density of the frames. On the other hand, we try to challenge MOT algorithms in terms of long term tracking by keeping sequences longer with less tracks.
The paper continues with the discussion of the related works in the following section that summarizes recent and popular approaches and datasets for MOT. Section \ref{SecProbDef} is devoted to the description of the MOT problem in terms of a surveillance perspective, discussing challenges and use cases. Section \ref{SecDataset} presents the details of the newly introduced dataset as well as the evaluation approach utilized that clearly analyzes the detection and tracking steps. Section \ref{SecExperiments} shows the experimental results that evaluate the performance of well-known MOT approaches in the proposed dataset. Section \ref{SecConclusion} concludes the paper with a discussion of the pros and cons of the MOT techniques as well as future comments.

\section{Related Work}
MOT is a common problem in most vision systems that enable relating objects among consecutive frames. This temporal relation provides an ID for each individual in a scene that is utilized to extend additional information about the attributes and behaviors of the objects and the scene statistics. In that manner, assigning an ID to an object and tracking it correctly are the first steps in gathering high-level inference about the scenes. In this section, we summarize the methods (in the scope of pedestrian tracking) proposed for MOT and the datasets that play a key role in the training and relatively high performance of modern CNNs.

\subsection{MOT Methods}
MOT techniques involve two main stages: detection and association, a.k.a., tracking. The detection determines the main purpose of the tracking by indicating the existence of an object type within a scene. Once objects are detected, next step is the association of objects. Throughout the MOT literature, various detection methodologies have been utilized involving moving object detection \cite{GMM}, blob detection \cite{BlobDetection}, feature detection \cite{SIFT}, predefined object detection \cite{HogHuman}. Until the last decade, hand-crafted features and rules have been utilized to detect objects in a scene. CNNs have dominated smart object detection with the help of a large amount of annotated objects and computational power, yielding wide application areas for machine learning. 

Once the objects are detected in a frame, association along the consecutive frames can be provided via two different approaches: assign a single object tracking per object or optimize a global cost function that relates all the objects detected in both frames. The first approach utilizes representation of the detected objects by defining a bounding box, foreground mask or sparse features. Then, these representations are searched along the next frame within a search region that is defined by the characteristics of the motion of the objects. In these types of methods, object detection is not required for each frame; instead, it is performed with low frequency to update the tracks with new observations. Feature matching \cite{SiftKLTtracker}, Kalman filters \cite{Kalman}, correlation-based matching \cite{KCFtracking} are the main tools for tracking representations among consecutive frames.

The second type of tracking exploits detection for each frame and relating the objects based on their similarities defined by several constraints such as position, shape, appearance, etc. The Joint Probabilistic Data Association Filter (JPDAF) \cite{JPDAF} and the Hungarian algorithm \cite{Hungarian} are the most widely used approaches to provide one-to-one matching along consecutive frames. In this way, independently extracted objects per frame are matched based on the similarity criteria. As CNNs improve, reID networks have also been utilized to yield robust similarities along objects in consecutive frames.     
The MOT literature has recently focused on the second approach, tracking-by-detection, where CNNs are utilized to detect objects \cite{SSD,YOLOv3,CenterNet} in each frame, and object similarities are extracted using different approaches \cite{sort,DeepSORT,yolov5-deepsort-osnet-2022,strongsort,bytetrack} that are fed into a matrix. The matrix representation involves the objects in the rows and columns, one side is devoted to track objects, the other is devoted to the new comers. Hungarian algorithm performed on the similarity matrix yields an optimum match of one to one. Methods are mostly differentiated according to the similarity formulation of the matrix, whereas the correspondence search is mostly achieved by a Hungarian algorithm.TransTrack \cite{transtrack}, TrackFormer \cite{TrackFormer}, and MOTR \cite{motr} have made attempts to use the attention mechanism in tracking objects in movies more recently with the current focus of utilizing transformers \cite{Transformers} in vision tasks.In these works, the query to associate the same objects across frames is transmitted to the following frames using the features of prior tracklets.To maintain tracklet consistency, the appearance information in the query is also crucial. 

\subsection{Datasets}
Datasets with ground-truth annotations are important for object detection and reID networks, which form the fundamental steps of modern MOT techniques. In this manner, we briefly summarize the existing person detection and multi-object tracking datasets which are within the scope of this study. In both sets, bounding boxes are utilized to define objects with a label that indicates the type of object, such as face, human, or vehicle. On the other hand, there are apparent differences between object detection and multi-object tracking datasets. Firstly, there is no temporal and spatial relationship between adjacent frames in object detection datasets. Second, there is no unique identification number of objects in object detection datasets. These differences make creation of multi-object tracking datasets more challenging than object detection datasets.

\begin{table}
  \centering
  \caption{A comparison of person detection datasets}
  \begin{adjustbox}{width=\linewidth}
  \begin{tabular}{@{}lccccccccc@{}}
    \toprule
     & Caltech \cite{CaltechPed} & KITTI \cite{KITTI} & COCOPersons \cite{COCOPerson} & CityPersons \cite{CityPersons} & CrowdHuman \cite{CrowdHuman} & VisDrone \cite{Visdrone} & EuroCityPersons \cite{EuroCity} & WiderPerson \cite{WiderPerson} & Panda \cite{Panda2020}\\
    \midrule
    \# countries & 1 & 1 & -& 3 & - & 1 & 12 & - & 1 \\
    \# cities & 1 & 1 & -&  27 & - & 1 & 31 & - & 1\\
    \# seasons & 1 & 1 & - & 3 & - & 1& 4 & 4 & 1 \\
    \# images (day/night) & 249884/- & 14999/- & 64115/- & 5000/-  & 15000/- & 10209/- & 40219/7118 & 8000/- & 45/- \\
    \# persons (day/night) & 289395/- & 9400/- & 257252 & 31514/- & 339565/- & 54200/- & 183182/35323 & 236073/- & 122100/-\\
    \# density(person/image) & 1 & 0.6 & 4 & 6.3 & 22.6 & 5.3 & 4.5 & 29.9  & 2713.8\\
    resolution & 640×480 & 1240×376 & - & 2048×1024 & - & 2000×1500 & 1920×1024 & 1400×800 & $>$ 25k×14k  \\
    weather & dry & dry & - & dry & - & - & dry,wet & dry,wet & - \\
    year & 2009 & 2013 & 2014 & 2017 & 2018 & 2018 & 2018 & 2019 & 2020 \\
    \bottomrule
  \end{tabular}
  \end{adjustbox}
  \label{detset}
\end{table} 

\subsection{Person detection datasets}
The widespread use of person detection datasets dates back to 2005 with the INRIA dataset \cite{INRIA}. Then after two more datasets emerge to serve the detection community in 2009 TudBrussels \cite{TudBrussels} and DAIMLER \cite{DAIMLER}. These three datasets increase structured progress of detection problem. However, as algorithms performance improves, these datasets are replaced by more diverse and dense datasets, e.g. Caltech \cite{CaltechPed} and KITTI \cite{KITTI}. CityPersons \cite{CityPersons} and EuroCityPersons \cite{EuroCity} datasets come to the fore with different country, city, climate and weather conditions. Despite the prevalence of these datasets, they all suffer from a low-density problem (person per frame); no more than 7. Crowd scenes are significantly underrepresented. CrowdHuman \cite{CrowdHuman} and WiderPerson \cite{WiderPerson} datasets attack this deficiency and increase the density to 22. Recently, the Panda \cite{Panda2020} dataset has been released, which is a very high resolution (25k x 15k) human-oriented detection and tracking dataset where relative object sizes are very small w.r.t. full image. This dataset focuses on very wide-angle surveillance through powerful processors by merging multiple high-resolution images. Another common surveillance dataset is Visdrone \cite{Visdrone}, which includes 11 different object types that differentiate humans and various vehicles. This dataset is captured by drones where view-points are much higher than surveillance, the platforms are moving and bird-eye views are observed. A summary of the datasets with various statistics is shown in Table~\ref{detset}.

\begin{table}
  \centering
  \caption{A comparison of \textit{SOMPT22} with other popular multi-object tracking datasets}
  \begin{adjustbox}{width=\linewidth}
  \begin{tabular}{@{}lcccccccccc@{}}
    \toprule
     & PETS \cite{PETS2009} & KITTI \cite{KITTI} & CUHK-SYSU \cite{CUHK-SYSU} & PRW \cite{PRW}  & PathTrack \cite{PathTrack} &  MOT17 \cite{MOT16} & MOT20 \cite{mot20} & BDD100K \cite{BDD100K} & DanceTrack \cite{DanceTrack} & \textit{SOMPT22}\\
    \midrule
    \# images & 795 & 8000 & 18184 & 6112 & 255000  &  11235  & 13410 & 318000 & 105855 & 21602 \\
    \# persons & 4476 & 47000 & 96143 & 19127 & 2552000 &  300373 & 1.6M & 3.3M & 1M & 801341\\
    \# density(ped/img) & 5.6 & 4.78 & 4.2 & 3.1 & 10 &  26 & 120 & 10.3 & 10 &   37\\
    \# tracks & 19 & 917 & 8432 & 450 & 16287 &  1666 & 3456 & 131000 & 990 &   997 \\
    camera angle & high view & eye-level & eye-level & eye-level & unstructured & eye-level & high view & eye-level & eye-level & high view \\
    fps  & 7 & - & - & - & 25 &  30 & 25 & 30 & 20 &  30\\
    camera state  & static & moving & moving & moving & moving &  moving/static & moving/static & moving/static & static &  static\\
    resolution & 768×576 & 1240×376 & 600×800 & 1920×1080 & 1280×720 &  1920×1080 & 1920×1080 & 1280×720 & 1920×1080 &  1920×1080\\
    year & 2009 & 2013 & 2016 & 2016 &  2017 & 2017 & 2020 & 2020 & 2021 &  2022\\
    \bottomrule
  \end{tabular}
  \end{adjustbox}
  \label{motset}
\end{table} 

\subsubsection{Multi-object tracking datasets} 
There is a corpus of multi-object tracking datasets that involve different scenarios for pedestrians. In terms of autonomous driving, the pioneering MOT benchmark is the KITTI suite \cite{KITTI} that provides labels for object detection and tracking in the form of bounding boxes. Visual surveillance-centric datasets focus on dense scenarios where people interact and often occluding each other and other objects. PETS \cite{PETS2009} is one of the first datasets in this application area. The MOTChallenge suite \cite{motchallenge} played a central role in the benchmark of multi-object tracking methods. This challenge provides consistently labeled crowded tracking sequences \cite{mot15,MOT16,mot20}. MOT20 \cite{mot20} boosts complexity of challenge by increasing density of the frames. MOT20 introduces a large amount of bounding boxes; however, the scenes are over-crowded and motion directions are that various to correspond real surveillance scenarios including squares and intersections. The recently published BDD100K \cite{BDD100K} dataset covers more than 100K videos with different environmental, weather, and geographic circumstances under unconstrained scenarios. In addition to these, the CUHK-SYSU \cite{CUHK-SYSU}, PRW \cite{PRW}, PathTrack \cite{PathTrack} and DanceTrack \cite{DanceTrack} datasets serve variety for multi-object tracking. These datasets are diverse in terms of static/moving camera; eye-level, high view angle and low/high resolution as shown in Table~\ref{motset}. 

\section{Problem Description}\label{SecProbDef}
As shown in Table~\ref{motset} the existing benchmarks mostly tackle MOT problem within autonomous driving perspective (eye level capture) due to the advances of vehicle technology. On the other hand, surveillance is one of the fundamental applications of video analysis that serves city-facility security, law enforcement, and smart city applications. As in most outdoor surveillance applications, the cameras are located high to cover large areas to observe and analyze. Stationary high-view cameras involve different content characteristics compared to common datasets provided in MOT, including severe projective geometry effects, larger area coverage, and longer but slower object motion. Therefore, it is beneficial to narrow the scope of MOT and analyze-optimize existing approaches for the surveillance problem. 

It is also important to consider the object type to be tracked during the scope limitation. In surveillance, the are two main object types under the spot, pedestrians and vehicles. Pedestrians have unpredictable motion patterns, interact in different ways with the other individuals, yielding various occlusion types, while showing 2D object characteristics due to thin structure. On the other hand, vehicles move faster along predefined roads with predictable (constant velocity-constant acceleration) motion models, interfere with other vehicles with certain rules defined by traffic enforcement, and suffer from object view point changes due to thickness in all 3 dimensions. In that manner, there are significant differences between the motion patterns and object view-point changes of pedestrians and vehicles that have influence on identification features and tracking constraints that definitely require careful attention. That is the main reason, the challenges differentiate by the type of objects as in \cite{AICITY21} and \cite{motchallenge}.

As a result of wide area coverage and high-view camera location, the viewpoints of pedestrians change significantly, where the distant objects can be observed frontally, and closer objects get high tilt angles w.r.t. camera. Besides, objects` slow relative motion yield longer tracks within the scene that requires tracking to be robust against various view changes for long periods of time. In this type of scene capture, object sizes in terms of image resolution get smaller, and the number of appearance-based tempters (human-like structures such as tree trunk, poles, seats, etc.) increases, while the videos involve less motion compared to eye level scene capture. Thus, object detection becomes more difficult and requires special attention to outliers as well as appearance changes in long tracks. In addition, consistent camera positioning enables the use of 3D geometry cues in terms of projective imaging, where several assumptions can be exploited such as objects closer to the camera being the occluders on a planar scene. Apart from the challenges in detection, especially observation of a longer object motion introduces new problems for tracking, mostly the change of floor illumination and object view point due to the object motion or occlusions. 

Constraining the MOT problem from a surveillance perspective, we propose a new annotated dataset and try to diagnose the capability of state-of-the-art MOT algorithms on pedestrians. As mentioned previously, MOT is achieved by two steps, detection and tracking where we also base our evaluation on the recent popular techniques for both steps. SOTA one-shot object detectors can be grouped into two: anchor-based, e.g., Yolov3 \cite{YOLOv3} and anchor-free, e.g., CenterNet \cite{CenterNet}. When we analyzed the 20 best performing MOT algorithms in the MOTChallenge \cite{motchallenge} benchmarks, we observed that these algorithms were based upon either CenterNet/FairMOT or Yolo algorithms. Therefore, we decided to build experiments around these basic algorithms to assess the success of detection and tracking in the \textit{SOMPT22} dataset. FairMOT \cite{FairMOT} and CenterTrack \cite{CenterTrack} are two one-shot multi-object trackers based on the CenterNet algorithm. FairMOT adds a reID head on top of the backbone to extract people embeddings. CenterTrack adds a displacement head to predict the next position of the centers of people. Two of the most common association methods are SORT \cite{sort} and DeepSORT \cite{DeepSORT}. SORT uses IOU(Intersection over union) and Kalman filter \cite{Kalman} as the criterion for linking detection to tracking. DeepSORT incorporates deep features of the detected candidates for linking detection to tracking. Three one-shot multi-object trackers (CenterTrack, FairMOT and Yolov5 \& SORT) and one two-stage multi-object tracker (Yolov5 \& DeepSORT) were trained to benchmark tracking performances.

\section{\textit{SOMPT22} Dataset}\label{SecDataset}
\subsection{Dataset Construction}
\noindent
\textbf{Video collection.} In order to obtain surveillance videos for MOT evaluation, 7/24 static cameras publicly streaming located at 6-8 m high poles are chosen all around the globe. Some of the chosen countries are Italy, Spain, Taiwan, the United States, and Romania. These cameras mostly observe squares and road intersections where pedestrians have multiple moving directions. Videos are recorded for around one minute at different times of the day to produce a variety of environmental conditions. In total, 14 videos were collected, by default using 9 videos as a training set and 5 as a test set. It is important to note that faces of the pedestrians are blurred to be anonymous in a way that does not affect pedestrian detection and reID features significantly. We conducted the object detection test with and without face blurring and have not observed any difference for base algorithms.

\noindent
\textbf{Annotation.} Intel's open-source annotation tool CVAT \cite{cvat} is used to annotate the collected videos. Annotation is achieved by first applying a pre-trained model to have rough detection and tracking labels, which are then fine tuned by human annotators. The annotated labels include bounding boxes and identifiers (unique track ID) of each person within MOTChallenge \cite{motchallenge} format. The file format is a CSV text file that contains one instance of an object per line. Each line contains information that includes \textit{frameID}, \textit{trackID}, \textit{top-left corner}, \textit{width and height}. In order to enable track continuity, partially and fully occluded objects are also annotated as long as they appear within the video again. Bounding boxes annotated with dimensions that overflow the screen size are trimmed to keep inside the image. Bounding boxes also include the part of the person who is occluded.

\begin{figure}[t!]
  \centering
  \includegraphics[width=4cm]{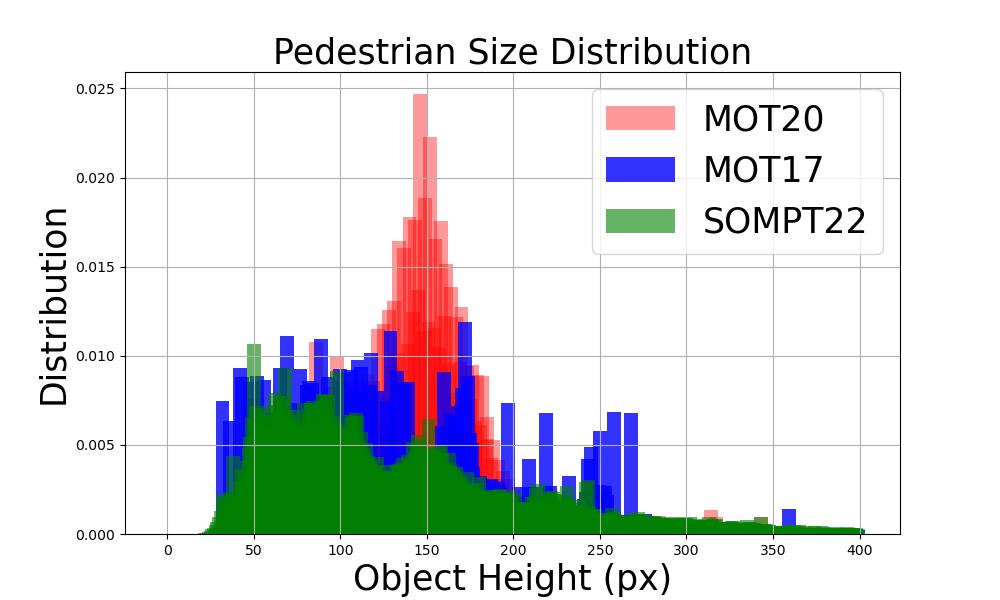}
  \includegraphics[width=4cm]{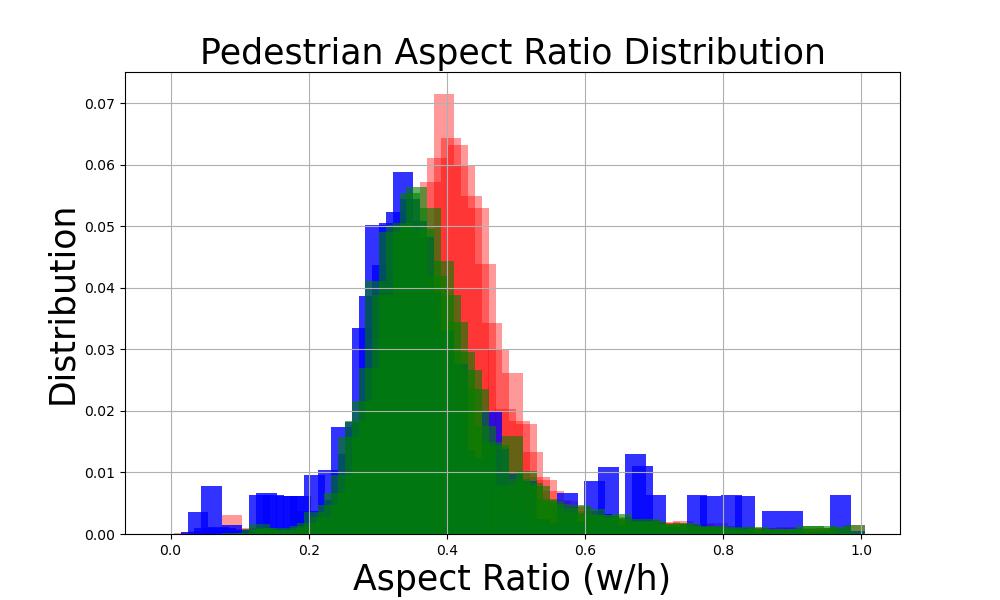}
  \includegraphics[width=4cm]{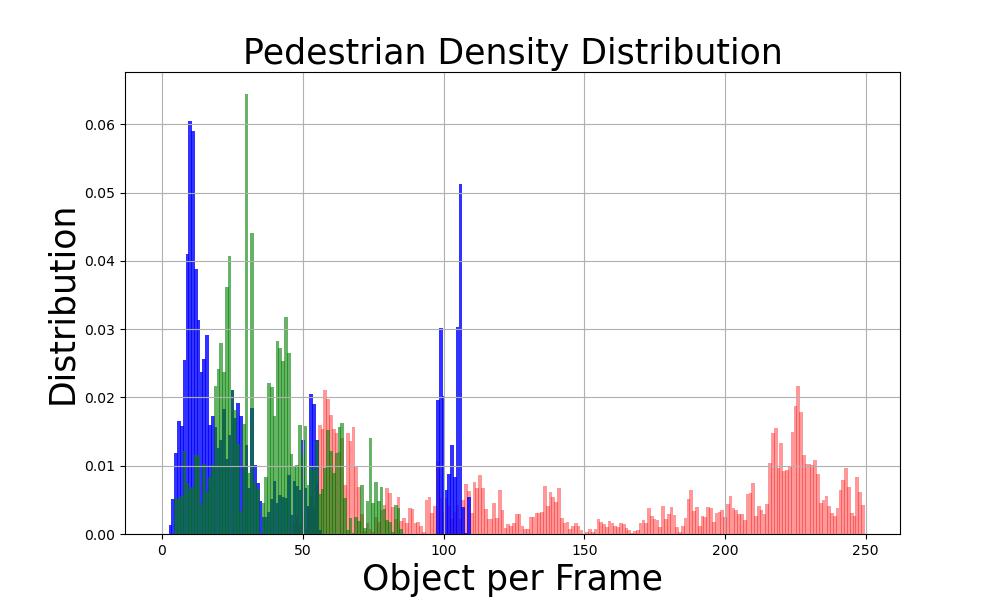}
  \includegraphics[width=4cm]{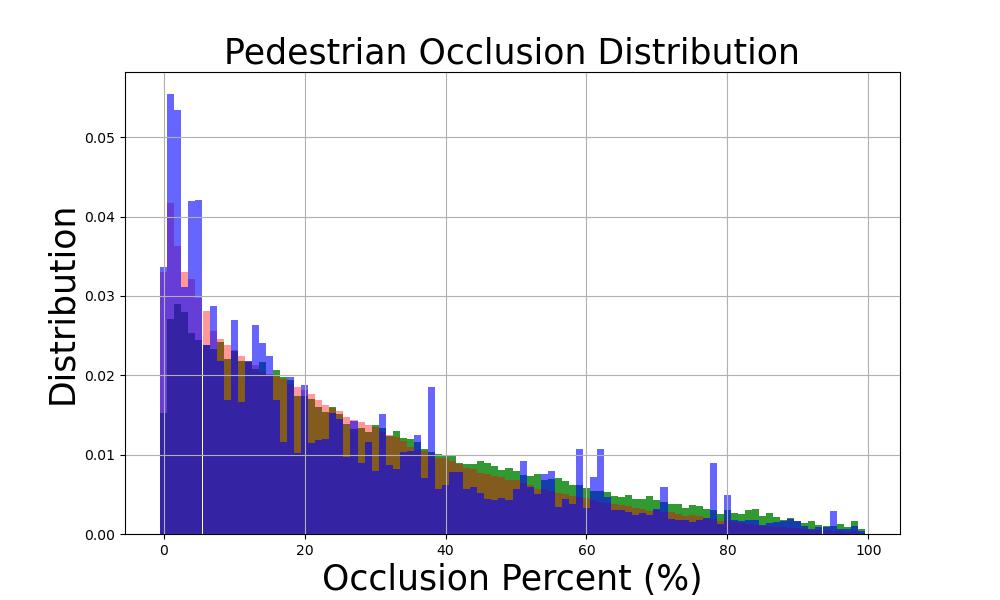}
  \includegraphics[width=4cm]{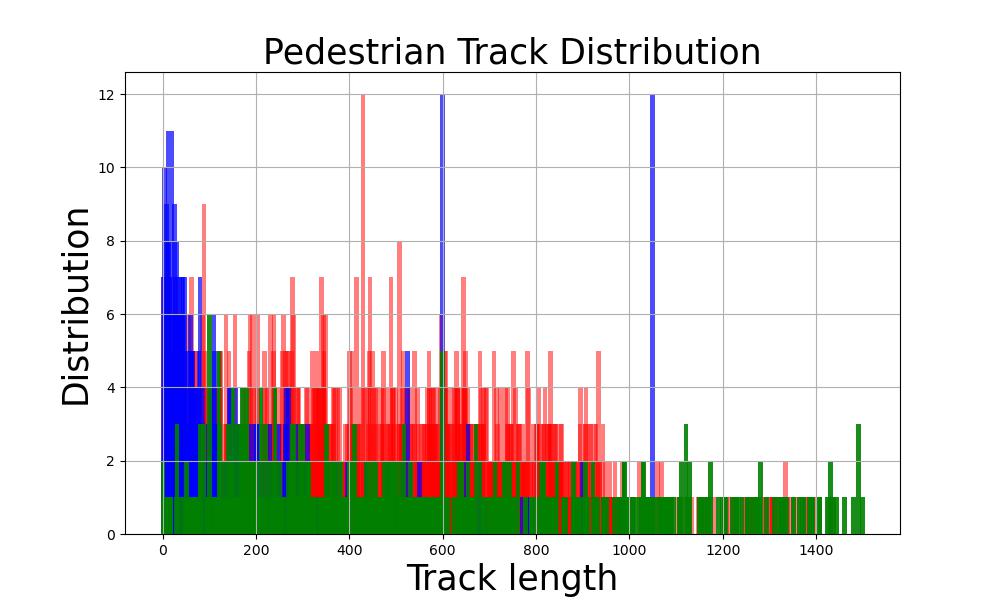}
  \caption{Statistics of MOT17, MOT20 and proposed \textit{SOMPT22} for persons of the training datasets (histogram of height, aspect ratio, density, occlusion and track length)}
  \label{fig:setDistribution}
\end{figure}

\subsection{Dataset Statistics}
Table~\ref{motset} presents some important statistics of the existing and proposed datasets. \textit{SOMPT22} has density of 37 pedestrians per frame that is between MOT17 and MOT20. MOT20 is a big step up in terms of number of person in MOTChallenge datasets. It is the most dense dataset right now. On the other hand, MOT17 and MOT20 are not mainly surveillance-oriented datasets rather they are proposed to challenge algorithms in terms of detection and occlusion. Especially in MOT20, videos are recorded during crowded events or at a metro station while people get off the train. The motion patterns of the pedestrians are not variable in these videos; each video includes a dominant direction with much fewer different directions. This is not a natural pattern of motion for people on surveillance cameras. On the contrary, people in \textit{SOMPT22} dataset act more spontaneously through city squares covering almost every direction. Figure \ref{fig:setDistribution} shows statistical benchmarking of MOT17, MOT20 and the proposed dataset. Although \textit{SOMPT22} has more images compared to MOT17 and MOT20, the number of tracks is the least. This shows that MOT17 and MOT20 have shorter tracklets with shorter sequences than \textit{SOMPT22}. A tracklet is a section of a track that a moving object is following as built by an image recognition system. This is an expected result, where surveillance oriented cameras cover larger field-of-views that enable longer observation of each individual. \textit{SOMPT22} provides a high-view dataset which is lacking in MOT datasets. In this manner, the \textit{SOMPT22} dataset challenges algorithms in terms of long-term detection, recognition, and tracking that require robust adaptation to changes in the scale and view points of pedestrians. We trained a YoloV5 \cite{YOLOv3} with a medium model backbone on the \textit{SOMPT22} training sequences, obtaining the detection results presented in Table \ref{tab:stat_det_seq}. A detailed breakdown of the detection bounding boxes on individual sequences and annotation statistics are shown in Table \ref{tab:stat}.

\begin{table}
  \centering
   \caption{\textit{SOMPT22} Dataset \& Detection Bounding Box Statistics}
  \begin{adjustbox}{width=\linewidth}
  \begin{tabular}{@{}lccccccc|cccc@{}}
    \toprule
    Annotation &  &  &  &  &  &  &  & Detection  & & \\
    \midrule
    Sequence & Resolution & Length(sec) & FPS & Frames & Boxes & Tracks & Density & nDet. & nDet/fr. & Min Height & Max Height \\
    \midrule
    Train &  &  &  &  &  &  &  & & & \\
    SOMPT22-02 & 1280x720 & 20 & 30 & 600 & 16021 & 45 & 28 & 15281 & 25 & 75 & 300 \\
    SOMPT22-04 & 1920x1080 & 60 & 30 & 1800 & 35853 & 44 & 20 & 35697 & 19 & 15 & 363 \\
    SOMPT22-05 & 1920x1080 & 60 & 30 & 1800 & 55857 & 40 & 31 & 58764 & 32 & 18 & 212 \\
    SOMPT22-07 & 1920x1080 & 60 & 30 & 1800 & 82525 & 94 & 46 & 81835 & 45 & 17 & 330 \\
    SOMPT22-08 & 1920x1080 & 60 & 30 & 1800 & 121708 & 138 & 68 & 124880 & 69 & 32 & 400 \\
    SOMPT22-10 & 1920x1080 & 60 & 30 & 1800 & 102077 & 139 & 57 & 104928 & 58 & 31 & 492\\
    SOMPT22-11 & 1920x1080 & 60 & 30 & 1800 & 41285 & 58 & 23 & 42706 & 23 & 22 & 417 \\
    SOMPT22-12 & 1920x1080 & 60 & 30 & 1800 & 72334 & 74 & 40 & 72987 & 40 & 16 & 506 \\
    SOMPT22-13 & 1920x1080 & 30 & 30 & 900 & 8244 & 38 & 9 & 8528 & 9 & 28 & 400\\
    \midrule
    Test &  &  &  &  &  &  &  & & &\\
    SOMPT22-01 & 1920x1080 & 60 & 30 & 1800 & 49970 & 75 & 28 & 59134 & 32 & 24 & 355\\
    SOMPT22-03 & 1280x720 & 40 & 30 & 1200 & 19527 & 42 & 16 & 17249 & 14 & 45 & 270\\
    SOMPT22-06 & 1920x1080 & 60 & 30 & 1800 & 122225 & 121 & 68 & 104148 & 57 & 29  & 321\\
    SOMPT22-09 & 1920x1080 & 60 & 30 & 1800 & 59092 & 57 & 33 & 50442 & 28 & 18  & 311 \\
    SOMPT22-14 & 1920x1080 & 30 & 30 & 902 & 14623 & 32 & 16 & 12781 & 14 & 29  & 439\\
    \midrule
    Total  &  & 12 mins &  & 21602 & 801341 & 997 & 37 & 789360 & 36 & 15 & 506\\
    \bottomrule
  \end{tabular}
  \end{adjustbox}
  \label{tab:stat}
\end{table}

\subsection{Evaluation Metrics}
The multi-object tracking community used MOTA \cite{CLEAR} as the main metric to benchmark for a long time. This measure combines three sources of error: false positives, missing targets, and identity switching. However, recent results have revealed that this metric places too much weight on detection rather than association quality, which owes too much weight to the quality of detection instead. Higher Order Tracking Accuracy (HOTA) \cite{HOTA} is proposed to correct this historical bias since then. HOTA is geometric mean of detection accuracy and association accuracy which is averaged across localization thresholds. In our benchmarking, we use HOTA as the main performance metric. We also use the AssA and IDF1 \cite{IDF1} scores to measure association performance. AssA is the Jaccard association index averaged over all matching detections and then averaged over localization thresholds. IDF1 is the ratio of correctly identified detections to the average number of ground-truth and computed detections. We use DetA and MOTA for detection quality. DetA is the Jaccard index of detection averaged above the localization thresholds. In this paper, we propose a new multi-pedestrian tracking dataset called \textit{SOMPT22}. This dataset contains surveillance oriented video sequences that captured on publicly streaming city cameras. The motivation is to reveal the bias in existing datasets that tend to be captured either on eye-level for autonomous driving systems or in high view and in crowded scenes. We believe that the ability to analyze the complex motion patterns of people in daily life on a well constrained surveillance scenario is necessary for building a more robust and intelligent trackers. \textit{SOMPT22} serves such a platform to encourage future work on this idea. localization thresholds. ID switch is the number of Identity Switches (ID switch ratio = \#ID switches / recall). The complexity of the algorithms is measured according to the processing cost (fps), including only the tracking step. The fps values may or may not be provided by the authors with non-standard hardware configurations. MOTChallenge \cite{motchallenge} does not officially take the reported fps of the algorithms into account in the evaluation process. 

\section{Experiments}\label{SecExperiments}
\subsection{Experiment Setup}
Table~\ref{tab:1} depicts the experimental configurations of the object detectors, multi-object trackers, and association algorithms. As can be seen in Table~\ref{detset}, CrowdHuman is a recent person detection dataset with a large volume and density of images. CenterTrack was pre-trained on CrowdHuman dataset \cite{CrowdHuman} by us. FairMOT and YoloV5 had already been pre-trained on it by the respective authors. The model parameters pre-trained on this dataset were utilized to initialize the detectors and trackers. Then, we fine-tune (transfer learning) CenterTrack, FairMOT and YoloV5 on the proposed \textit{SOMPT22} train dataset via 240, 90 and 90 epochs correspondingly. For the sake of fairness, we keep the network input resolution fixed for all detectors and trackers during the training and inference phase. We followed the presented training protocols of the detectors and trackers in their respective source codes. Therefore, each object detector and tracker has its own pre-processing techniques, data augmentation procedures, hyper-parameter tuning processes as well as accepted dataset annotation formats e.g. yolo \cite{YOLOv3} for YoloV5, MOTChallenge \cite{motchallenge} for FairMOT and COCO \cite{COCOPerson} for CenterTrack. DeepSORT algorithm has CNN based feature extractor module. This module is pre-trained on the Market1501 \cite{Market1501} public reID dataset. All algorithms are implemented and executed on the PyTorch framework with Python, some of which are provided by the corresponding authors. The inference experiments were conducted on an Intel i7-8700k CPU PC with Nvidia GTX1080ti (11GB) GPU. 

As following tracking-by-detection technique, detection is performed for each frame independently. We use the Kalman filter \cite{Kalman} and bounding box intersection over union as initial stages for all trackers to associate the detection results along consecutive frames. Further details of the experimented trackers are given in Table~\ref{tab:3}. YoloV5 object detector constructs a two-stage multi-object tracker in collaboration with the DeepSORT algorithm. YoloV5 \& SORT, CenterTrack and FairMOT algorithms are three one-shot trackers that have only one backbone to extract deep features from objects. The multi-object tracker formed by cascading YoloV5 and SORT algorithms is classified as one-shot tracker due to the fact that association is performed purely on CPU. These three multi-object trackers are trained in end-to-end fashion. DeepSORT association algorithm and FairMOT benefit from reID features, while CenterTrack and SORT only accomplish the association task without reID features. CenterTrack exploits an additional head within detection framework that provides prediction of the displacement.

\begin{table}
  \centering
  \caption{Algorithms \& Specs (* not trained by us)}
  \begin{adjustbox}{width=\linewidth}
  \begin{tabular}{@{}lcccccc@{}}
    \toprule
    Algorithms & Function & Type  & Backbone & Resolution & Train Dataset \\
    \midrule
    CenterTrack \cite{CenterTrack} & Multi-object tracker & One-shot & DLA34 & 640x640 & CH / + \textit{SOMPT22} \\
    FairMOT \cite{FairMOT} & Multi-object tracker & One-shot & DLA34 & 640x640 & CH* / + \textit{SOMPT22}\\
    YoloV5 \cite{YOLOv3} & Object Detector & One-shot & Darknet & 640x640 & CH* / + \textit{SOMPT22}\\
    DeepSORT \cite{DeepSORT} & Associator & - & CNN & 64x128 & Market1501* \cite{Market1501} \\
    SORT \cite{sort} & Associator & - & - & 64x128 & - \\
    \bottomrule
  \end{tabular}
  \end{adjustbox}
  \label{tab:1}
\end{table}

\begin{table}
  \caption{Association Methods of MOT Algorithms}
  \centering
  \begin{adjustbox}{width=\linewidth}
  \begin{tabular}{@{}lccccc@{}}
    \toprule
    Methods & Box IOU & Re-ID Features & Kalman Filter &  Displacement\\
    \midrule
    CenterTrack \cite{CenterTrack} & \checkmark & &  \checkmark & \checkmark\\
    FairMOT \cite{FairMOT} & \checkmark & \checkmark & \checkmark & \\
    YoloV5 \& DeepSORT \cite{yolov5-deepsort-osnet-2022} & \checkmark &\checkmark & \checkmark & \\
    YoloV5 \cite{YOLOv3} \& SORT \cite{sort} & \checkmark &  & \checkmark & \\
    \bottomrule
  \end{tabular}
  \end{adjustbox}
  \label{tab:3}
\end{table}

\subsection{Benchmark Results}
In this section, we will compare and contrast the performance of the four aforementioned trackers according to the HOTA \cite{HOTA} and CLEAR \cite{CLEAR} metrics and the inference speed, as shown in Table~\ref{tab:4}. We can observe that the detection performance (DetA) of CenterTrack is better than that of FairMOT, which may be due to the displacement head that improves the localization of people. On the other hand, the association performance (AssA) of FairMOT is better than that of CenterTrack with the help of the reID head that adds a strong cue to the association process. CenterTrack requires less computation sources compared to FairMOT. According to the HOTA score which is the geometric mean of DetA and AssA, combination of YoloV5 and SORT variants outperform the other techniques significantly. The key role in this result belongs to the detection accuracy, where YoloV5 improves the detection by at least 10\%. DeepSORT and SORT approaches perform on similar to each other with certain expected deviations. DeepSORT adds reID based matching of object patches over SORT and decreases ID switch by 80\% while increasing the computational complexity by x2.5. However, the use of reID representations introduces some decrease in association accuracy (AssA), which is probably due to long tracklets that change appearance significantly. YoloV5 is an anchor-based object detector. Fine selection of anchor combination seems to lead better detection performance on surveillance cameras than methods that utilize anchor-free methods e.g. CenterTrack and FairMOT. This reveals the importance of detection during the track-by-detect paradigm that is the most common approach in MOT literature. MOT20 is the most similar public dataset to \textit{SOMPT22} in the literature in terms of camera perspective. Therefore, we repeated same experiments to observe contribution of \textit{SOMPT22} to MOT algorithms on MOT20 dataset as well. Table~\ref{tab:5} shows the benchmark of the MOT algorithms in the MOT20 train set. The comparative results are parallel to those given in Table~\ref{tab:4}, where the YoloV5 and (Deep)SORT methods outperform. It is also clear that transfer learning in our proposed dataset \textit{SOMPT22} increases performance. 

\begin{table}
  \centering
  \caption{MOT algorithms performance on \textit{SOMPT22} Test Set before and after fine-tuning on \textit{SOMPT22} ($\uparrow$: higher is better, $\downarrow$: lower is better)}
  \begin{adjustbox}{width=\linewidth}
  \begin{tabular}{@{}lcccccccc@{}}
    \toprule
    Methods & HOTA $\uparrow$ &  DetA $\uparrow$ & AssA $\uparrow$ & MOTA $\uparrow$ & IDF1 $\uparrow$ & IDsw $\downarrow$  & FPS $\uparrow$  \\
    \midrule
    CenterTrack \cite{CenterTrack} & 23.4 / 32.9 & 31.2 / 42.3 & 18.0 / 26.1 & 35.6 / 43.5 & 23.4 / 35.3 & 5426 / 3843  & 57.7 \\
    FairMOT \cite{FairMOT} & 36.3 / 37.7 & 38.2 / 34.6 & 35.5 / 41.8 & 41.0 / 43.8 & 44.8 / 42.9 & 1789 / 1350  & 12\\    
    YoloV5 \& DeepSORT \cite{yolov5-deepsort-osnet-2022} & 39.5 / 43.2 & 48.8 / \textbf{47.2} & 32.4 / 39.8 & 58.4/ 55.9 & 47.1 / 53.3 & 162 / \textbf{152}  & 26.4\\               
    YoloV5 \cite{YOLOv3} \& SORT \cite{sort} & 44.5 / \textbf{45.1} & 47.8 / 47.1 & 41.7 / \textbf{43.5} & 59.4 / \textbf{56} & 55.9 / \textbf{55.1} & 747 / 822  & \textbf{70.2}\\
    \bottomrule
  \end{tabular}
  \end{adjustbox}
  \label{tab:4}
\end{table}

\begin{table}
  \centering
  \caption{MOT algorithms performance on MOT20 Train Set before and after fine-tuning on \textit{SOMPT22} ($\uparrow$: higher is better, $\downarrow$: lower is better) }
  \begin{adjustbox}{width=\linewidth}
  \begin{tabular}{lccccccccc}
    \toprule
    Methods  & HOTA $\uparrow$ &  DetA $\uparrow$ & AssA $\uparrow$ & MOTA $\uparrow$ & IDF1 $\uparrow$ & IDsw $\downarrow$ \\
    \midrule
     CenterTrack \cite{CenterTrack}                      & 13.2 / 18.8 & 12.7 / 28.8  & 14.2 / 12.5  & 14.3 / 33.4  & 12.1 / 20.7 & 64 / 1318\\
     FairMOT \cite{FairMOT}                              & 15.4 / 18.1 & 7.9 / 16.8  & 30.9 / 22.8  &  3.3 / -21.1 & 11.4 / 15.1 & 232 / \textbf{153}\\         
     YoloV5 \& DeepSORT \cite{yolov5-deepsort-osnet-2022}& 26.8 / 41.5 & 35.8 / \textbf{52.5}  & 20.4 / 33.0  & 47.3 / \textbf{67.5}  & 33.2 / 51.0 & 345 / 5859 \\                           
    YoloV5 \cite{YOLOv3} \& SORT \cite{sort}              & 32.1 / \textbf{48.0}  & 32.7 / 50.4  & 31.7 / \textbf{45.9}  & 42.7 / 63.2  & 44.1  / \textbf{62.6} & 4788 / 3013 \\                              
    \bottomrule
  \end{tabular}
  \end{adjustbox}
  \label{tab:5}
\end{table}

As mentioned, combination of SORT based associator and anchor-based object detector YoloV5 performs better than one-shot MOT algorithms. In addition, detection performance is critical for overall tracking. To observe the contribution of the \textit{SOMPT22} dataset to the performance of object detection solely, we evaluated the YoloV5 object detector on the test set of \textit{SOMPT22} after fine tuning with the same dataset. Detection scores are shown in Table \ref{tab:yolov5det}. The precision and recall measurements are calculated as 0.89 and 0.68 respectively, which indicates that there is still room for improvement of detection under surveillance scenario. The problem in surveillance is the wide field of view that results in small objects which are rather difficult to detect. We provide these public detections as a baseline for the tracking challenge so that the trackers can be trained and tested. 

\begin{table}
  \centering
   \caption{Overview of Performance of YoloV5 detector trained on the \textit{SOMPT22} Training Dataset}
  \begin{adjustbox}{width=6.5cm}
  \begin{tabular}{@{}lccccc@{}}
    \toprule
    Sequence & Precision & Recall  & AP@.5 & AP@.5:.95 \\
    \midrule
    Train &  &  &  &  &    \\
    SOMPT22-02 & 0.98 & 0.89 & 0.96 & 0.61   \\
    SOMPT22-04 & 0.98 & 0.94 & 0.98 & 0.73   \\
    SOMPT22-05 & 0.97 & 0.92 & 0.97 & 0.73   \\
    SOMPT22-07 & 0.96 & 0.85 & 0.93 & 0.66   \\
    SOMPT22-08 & 0.96 & 0.82 & 0.93 & 0.66   \\
    SOMPT22-10 & 0.97 & 0.84 & 0.94 & 0.68   \\
    SOMPT22-11 & 0.98 & 0.92 & 0.97 & 0.74   \\
    SOMPT22-12 & 0.99 & 0.95 & 0.98 & 0.83   \\
    SOMPT22-13 & 0.98 & 0.93 & 0.95 & 0.67   \\
    \midrule
    Test &  &  &  &  &    \\
    SOMPT22-01 & 0.85 & 0.72 & 0.79 & 0.41   \\
    SOMPT22-03 & 0.95 & 0.72 & 0.80 & 0.42   \\
    SOMPT22-06 & 0.89 & 0.66 & 0.77 & 0.43   \\
    SOMPT22-09 & 0.87 & 0.51 & 0.60 & 0.33   \\
    SOMPT22-14 & 0.94 & 0.70 & 0.80 & 0.50   \\
    \bottomrule
  \end{tabular}
  \end{adjustbox}
  \label{tab:stat_det_seq}
\end{table}

\begin{table}
  \centering
  \caption{The performance of YoloV5 \cite{YOLOv3} object detector evaluated on the test set of \textit{SOMPT22}}
  \begin{tabular}{@{}lccccc@{}}
    \toprule   
    Train Set & Precision & Recall  & AP@.5 & AP@.5:.95\\
    \midrule
    CH/+\textit{SOMPT22}  & 0.88/0.89 & 0.65/0.68 & 0.74/0.79 &  0.40/0.42 \\
    \bottomrule
  \end{tabular}
  \label{tab:yolov5det}
\end{table}

\begin{figure}[t!]
  \includegraphics[width=\textwidth]{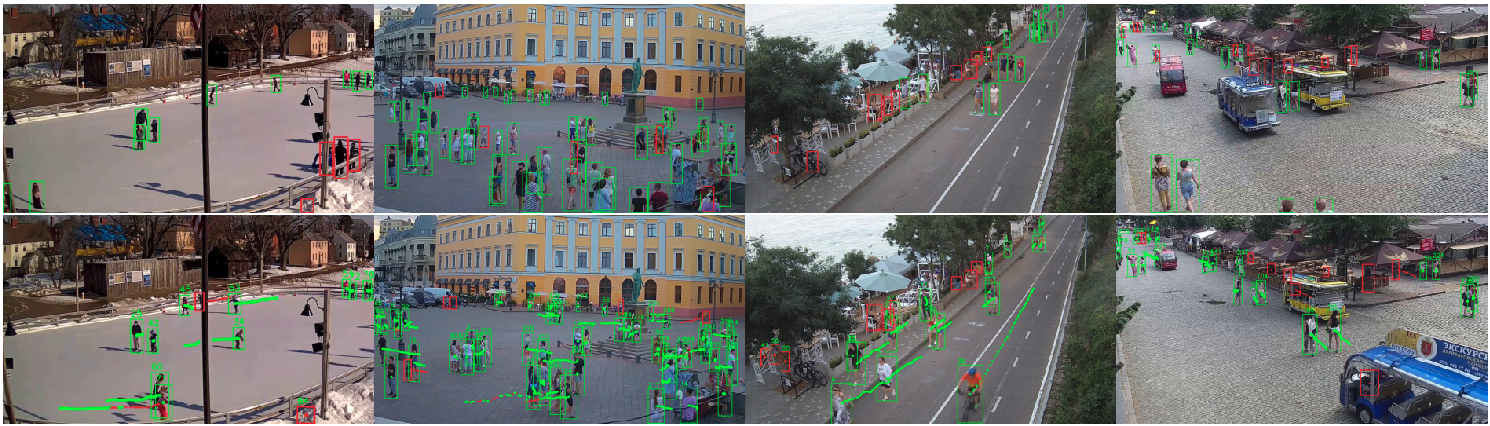}\hfill
  \caption{Some success/failure cases of YoloV5 \& SORT MOT algorithm on \textit{SOMPT22}. First row shows true detections in green and false detections in red. Second row shows true trajectories in green and fragmented trajectories in red}
  \label{fig:suc-fail}
\end{figure}

Figure~\ref{fig:suc-fail} shows some success and failure cases of YoloV5 \& SORT method on \textit{SOMPT22} dataset where green color indicates successful detection and tracks and red vice-versa. The detectors fail along cluttered regions where pedestrians are occluded or pedestrian-like structures show off, on the other hand, trackers fail. Detector failure leads to ID switch, trajectory fragmentation and losing long-term tracking. The experimental results on the proposed dataset indicate that SOTA is still not very efficient. On the other hand, detection plays a key role in the overall tracking performance. Thus, enrichment of detectors with additional attributes and representations that support association of objects seem be a competitive alternative in terms of lower computational complexity and higher performance. Two-stage approaches perform obviously better and provide room for improvement such that they are fast and enable a low number of ID switches with a much more accurate association.

\section{Conclusion \& Future Work} \label{SecConclusion}
In this paper, we propose a new multi-pedestrian tracking dataset: \textit{SOMPT22}. This dataset contains surveillance oriented video sequences that captured on publicly streaming city cameras. The motivation is to reveal the bias in existing datasets that tend to be captured either on eye-level for autonomous driving systems or in high view and in crowded scenes. We believe that the ability to analyze the complex motion patterns of people in daily life on a well constrained surveillance scenario is necessary for building a more robust and intelligent trackers. \textit{SOMPT22} serves such a platform to encourage future work on this idea.

The benchmark of four most common tracking approaches in \textit{SOMPT22} demonstrates that the multi-object tracking problem is still far from being solved with at most 48 \% HOTA score and requires fundamental modifications prior to usage of heuristics in specific scenarios.

FairMOT and CenterTrack multi-object trackers show complementary performance for the detection and association part of the tracking task. On the other hand, improved detection with YoloV5 followed by SORT based trackers outperform joint trackers. Moreover, SORT provides higher tracking scores, apart from ID switch compared to DeepSORT. These indicate that detection is the key to better tracking, and reID features require special attention to be incorporated within the SORT framework.

\clearpage
%
%

\bibliographystyle{splncs04}
\bibliography{rws2022}
\end{document}